\documentclass[runningheads]{llncs}
\usepackage{graphicx}
\usepackage{booktabs} 

\usepackage{multirow}
\usepackage{amsmath}

\DeclareMathOperator*{\argmin}{arg\,min}

\usepackage[ruled]{algorithm2e} 

\SetAlFnt{\small}
\SetAlCapFnt{\small}
\SetAlCapNameFnt{\small}
\SetAlCapHSkip{0pt}
\IncMargin{-\parindent}

\begin{document}
\title{Data-Driven Optimization of Public Transit Schedule}

%
%
\author{Sanchita Basak, Fangzhou Sun, Saptarshi Sengupta \and
Abhishek Dubey}
\authorrunning{Basak, Sun, Sengupta and Dubey}
%
\institute{Department of EECS, Vanderbilt University, Nashville, TN, USA \email{\{sanchita.basak@vanderbilt.edu, fzsun316@gmail.com, saptarshi.sengupta@vanderbilt.edu, abhishek.dubey@vanderbilt.edu\}} }

\maketitle              
\begin{abstract}
Bus transit systems are the backbone of public transportation in the United States. An important indicator of the quality of service in such infrastructures is on-time performance at stops, with published transit schedules playing an integral role governing the level of success of the service. However there are relatively few optimization architectures leveraging stochastic search that focus on optimizing bus timetables with the objective of maximizing probability of bus arrivals at timepoints with delays within desired on-time ranges. In addition to this, there is a lack of substantial research considering monthly and seasonal variations of delay patterns integrated with such optimization strategies. To address these, this paper makes the following contributions to the corpus of studies on transit on-time performance optimization: (a) an unsupervised clustering mechanism is presented which groups months with similar seasonal delay patterns, (b) the problem is formulated as a single-objective optimization task and a greedy algorithm, a genetic algorithm (GA) as well as a particle swarm optimization (PSO) algorithm are employed to solve it, (c) a detailed discussion on empirical results comparing the algorithms are provided and sensitivity analysis on hyper-parameters of the heuristics are presented along with execution times, which will help practitioners looking at similar problems. The analyses conducted are insightful in the local context of improving public transit scheduling in the Nashville metro region as well as informative from a global perspective as an elaborate case study which builds upon the growing corpus of empirical studies using nature-inspired approaches to transit schedule optimization.

\end{abstract}
%
%
%
\section{Introduction}

Bus systems are the backbone of public transportation in the US, carrying over 47\% of all public passenger trips and 19,380 million passenger miles in the US \cite{neff20172016} . For the majority of cities in the US which do not have enough urban forms or budget to build expensive transit infrastructures like subways, the reliance is on buses as the most important transit system since bus systems have advantages of relatively low cost and large capacity. Nonetheless, the bus system is also one of the most unpredictable transit modes. Our study found that the average on-time performance across all routes of Nashville bus system was only 57.79\% (see Section~\ref{opt18:sec:sensitivity}). The unpredictability of delay has been selected as the top reason why people avoid bus systems in many cities \cite{apta2017report}. 

Providing reliable transit service is a critical but difficult task for all metropolis in the world. To evaluate service reliability, transit agencies have developed various indicators to quantify public transit systems through several key performance measurements from different perspectives \cite{benn1995bus}. In the past, a number of technological and sociological solutions have helped to evaluate and reduce bus delay.  Common indicators of public transit system evaluation include schedule adherence, on-time performance, total trip travel time, etc. In order to track the transit service status, transit agencies have installed AVL on buses to track their real-time locations. However, the accuracy of AVL in urban areas is quite limited due to the low sampling rate (every minute) and the impact of high buildings on GPS devices. To have some basic controls during bus operation, public transit agencies often use time point strategies, where special timing bus stops (time points are special public transit stops where transit vehicles try to reach at scheduled times) are deployed in the middle of bus routes to provide better arrival and departure time synchronizations.

An effective approach for improving bus on-time performance is creating timetables that maximize the probability of on-time arrivals by examining the actual delay patterns. When designing schedules for real-world transport systems (e.g. buses, trains, container ships or airlines), transport planners typically adopt a tactical-planning approach \cite{fan2006optimal}.  Conventionally, metro transit engineers analyze the historical data and adjust the scheduled time from past experience, which is time consuming and error prone. A number of studies have been conducted to improve bus on-time performance by reliable and automatic timetabling. Since the timetable scheduling problem is recognized to be an NP-hard problem \cite{wu2016multi}
, many researchers have employed heuristic algorithms to solve the problem. The most popular solutions include ad-hoc heuristic searching algorithms (e.g. greedy algorithms), neighborhood search (e.g. simulated annealing (SA) and tabu search (TS)), evolutionary search (e.g. genetic algorithm) and hybrid search \cite{szeto2011simultaneous}.

\begin{figure*}[t]
\centering
        \includegraphics[scale=0.3]{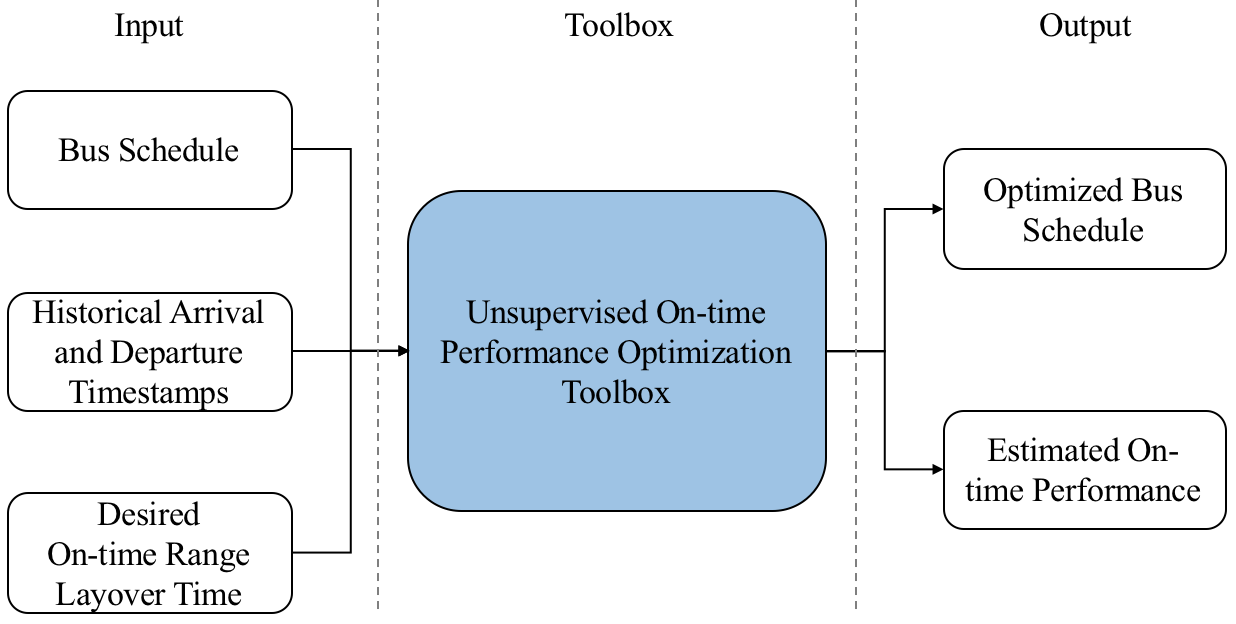}
    
\caption{The proposed toolbox for bus on-time performance optimization. City planners use bus schedule, historical trip information and desired on-time range and layover time, and get outputs of optimized timetable as well as estimated on-time performance.}
\label{opt18:fig:toolbox}
\end{figure*}

However, there are few stochastic optimization models that focus on optimizing bus timetables with the objective of maximizing the probability of bus arrivals at timepoint with delay within a desired on-time range (e.g. one minute early and five minutes late), which is widely used as a key indicator of bus service quality in the US \cite{arhin2014bus}.  A timepoint is a bus stop that is designed to accurately record the timestamps when buses arrive and leave the stop. Bus drivers use timepoints to synchronize with the scheduled time.  For example, to quantify bus on-time arrival performance, many regional transit agencies use the range of [-1,+5] minutes compared to the scheduled bus stop time as the on-time standard to evaluate bus performance using historical data \cite{arhin2014bus}. The actual operation of bus systems is vulnerable to many internal and external factors. The external factors include urban events (e.g., concerts, sporting events, etc.), severe weather conditions, road construction, passenger and bicycle loading/offloading, etc. One of the most common internal factors is the delay between two consecutive bus trips, where the arrival delay of previous trips causes departure delay of the next trip. Furthermore, there are monthly and seasonal variation in the actual delay patterns, but most transit agencies publish a uniform timetable for the next several months despite the variations. How to cluster the patterns and optimize timetables separately remains an open problem. Furthermore, heuristic optimization techniques have attracted considerable attention, but finding the optimal values of hyper-parameters are difficult, since they depend on nature of problem and the specific implementation of the heuristic algorithms, and are generally problem specific. 



\hspace{1 em}

\textbf{Research Contributions:} In this paper, the monthly and seasonal delay patterns are studied and outlier analysis and clustering analysis on bus travel times to group months with similar patterns together are carried out. The feature vectors are aggregated by routes, trips, directions, timepoint segments and months encompassing mean, median and standard deviation of the historical travel times. This work significantly extends prior work on the problem as in \cite{sun2017unsupervised}. 
Along with a greedy algorithm and a Genetic Algorithm (GA), swarm based optimization algorithm has been introduced in this work, where the semi-autonomous agents in the swarm can update their status guided by the full knowledge of the entire population state. Thus, Particle Swarm Optimization (PSO) \cite{488968} algorithm is employed in this work to generate new timetables for month clusters that share similar delay patterns with the goal of testing both evolutionary computing and swarm intelligence approaches. It is observed that the optimized on-time performance averaged across all bus routes has increased by employing PSO as compared to that by GA. Also  the execution times of PSO are much less than GA and are more stable indicating lesser variability of results over different runs. Sensitivity analysis on choosing the optimal hyper-parameters for the proposed heuristic optimization algorithms are also presented. A stability analysis of the respective algorithms have been put forward by studying the on-time performance and execution time over several runs. The overall workflow of the proposed optimization mechanisms is illustrated in Figure~\ref{opt18:fig:toolbox}. 

\hspace{1 em}




The rest of the paper is organized as follows:
Section~\ref{opt18:sec:relatedwork} compares our work with related work on transit timetabling;
Section~\ref{opt18:sec:formulation} presents the problem formulation;
Section~\ref{opt18:sec:data} presents the details of the transit data stores;
Section~\ref{opt18:sec:solutions} discusses the timetable optimization mechanisms used;
Section~\ref{opt18:sec:evaluations} evaluates the performance of the optimization mechanisms and presents sensitivity analysis results;
Section~\ref{opt18:sec:conclusion} presents concluding remarks and future work.

\section{Related Work and Challenges}
\label{opt18:sec:relatedwork}
This section compares our system with related work on transit timetable scheduling. A number of studies have been conducted to provide timetabling strategies for various objectives: (1) minimizing average waiting time \cite{wang2017data} (2) minimizing transfer time and cost \cite{chakroborty1995optimal}\cite{hairong2009optimal}\cite{szeto2011simultaneous}, (3) minimizing total travel time \cite{nayeem2014transit}, (4) maximizing number of simultaneous bus arrivals \cite{eranki2004model}, \cite{ibarra2012synchronization}, (5) minimizing the cost of transit operation \cite{ting2005schedule}, (6) minimizing a mix of cost (both the user's and the operator's) \cite{chakroborty2003genetic}.

The design of timetable with maximal synchronizations of bus routes without bus bunching has been researched by Ibarra-Rojas et al. \cite{ibarra2012synchronization}. The bus synchronization strategy has been discussed from the perspective of taking waiting time into account in the transfer stops in the work of Eranki et al. \cite{eranki2004model}. An improved GA in minimizing passenger transfer time considering traffic demands has been explored by Yang et al. \cite{hairong2009optimal}. Traffic and commuter demand has also been considered in the work by Wang et al. \cite{wang2017data}. Other than employing optimization algorithms several deep learning techniques \cite{DBLP:journals/corr/abs-1905-13294} have been applied in bus scheduling problems \cite{10.1007/978-3-319-31753-3_44}.

Nayeem et al. \cite{nayeem2014transit} set up the optimization problem over several criteria, such as minimizing travel time and number of transfers and maximizing passenger satisfaction. A route design and neighborhood search through genetic algorithm minimizing number of transfers has been discussed by Szeto et al. \cite{szeto2011simultaneous}. Zhong et al. \cite{psotraffic} used improved Particle Swarm Optimization for recognizing bus rapid transit routes optimized in order to serve maximum number of passengers.

\subsection{Research Challenges}


\textbf{\hspace{1 em} (a) Clustering Monthly and Seasonal Variations in Historical Arrival Data:} Studying the historical travel time at segments can be an effective way to set bus timetables. However, existing work doesn't consider the monthly and seasonal variation in historical monthly data, and the variation can be utilized for better scheduling. Generating one timetable for all months may not be the best solution. As traffic and delay patterns are prone to changes over seasonal variations and various times, we generate clusters grouping months with unsupervised algorithm and develop optimization strategies for the generated clusters. We evaluate the proposed mechanism via simulation. The cluster-specific schedule is shown to further increase the on-time performance compared to generating one uniform timetable. 

\vspace{1 em}
\textbf{(b) Computing Efficiently and Accurately in the Solution Space:} Transit performance optimization techniques rely on historical delay data to set up new timetables. However, the large amount of historical data makes it a challenge to compute efficiently. For example, Nashville MTA updates the bus schedule every 6 months but each time there are about 160,000 historical record entries to use. Moreover, the solution space has typically very large under constraints (e.g., sufficient dwelling time at bus stops, adequate layover time between trips, etc.).
A suitable optimization algorithm is necessary for efficient and accurate computation. Since this is a discrete-variable optimization problem, gradient-based methods cannot be used and gradient-free methods need to be considered. A naive algorithm for discrete optimization is exhaustive search, i.e., every feasible time is evaluated and the optimum is chosen. Exhaustive search works for a small finite number of choices, and cannot be used for high-dimensional problems. Genetic algorithm \cite{chakroborty1995optimal}\cite{chakroborty2003genetic}, as well as particle swarm optimization \cite{488968} are used commonly in solving heuristic problems . Thus we consider applying genetic algorithm and particle swarm optimization (PSO) in the context. Section~\ref{opt18:sec:solutions} describes the key steps of how we apply greedy, genetic and PSO algorithms to solve the timetable optimization problem.

\section{Problem Formulation}
\label{opt18:sec:formulation}


\begin{table}[tb]
\centering
\caption{The scheduled time and recorded actual arrival and departure time of two sequential trips that use the same bus of route 4 on Aug. 8, 2016. The arrival delay at the last timepoint of the first trip accumulates at the first timepoint of the second trip.}
\label{opt18:tbl:example}
\begin{tabular}{|l|l|l|l|l|l|}
\hline
                              &                       & \multicolumn{4}{c|}{Timepoints}           \\ \hline
                              &                       & MCC4\_14 & SY19     & PRGD     & GRFSTATO \\ \hline \hline
\multirow{3}{*}{Trip 1} & Scheduled Time        & 10:50 AM & 11:02 AM & 11:09 AM & 11:18 AM \\ \cline{2-6} 
                              & Actual Arrival Time   & 10:36 AM & 11:10 AM & 11:18 AM & 11:27 AM \\ \cline{2-6} 
                              & Actual Departure Time & 10:50 AM & 11:10 AM & 11:18 AM & 11:30 AM \\ \hline \hline
\multirow{3}{*}{Trip 2} & Scheduled Time        & 11:57 AM & 11:40 AM & 11:25 AM & 11:20 AM \\ \cline{2-6} 
                              & Actual Arrival Time   & 12:11 PM & 11:51 AM & 11:34 AM & 11:27 AM \\ \cline{2-6} 
                              & Actual Departure Time & 12:11 PM & 11:51 AM & 11:34 AM & 11:30 AM \\ \hline
\end{tabular}
\end{table}

Typically, transit delay are not only affected by external factors (such as traffic, weather, travel demand, etc.), but also by some internal factors. For example, the accumulated delay occurred on previous trips may cause a delay in consecutive trips by affecting the initial departure time of the next trip. 
In order to illustrate the problem context with simplicity and without generality, we take two sequential bus trips of route 4 in Nashville as an example (the scheduled time and the actual arrival and departure time recorded on Aug. 8, 2016 are shown in Table~\ref{opt18:tbl:example}) to describe the optimization problem. 
On each service day, after a vehicle of the first trip (121359) arrives at the last stop (Timepoint GRFSTATO) with scheduled time of 11:18 AM, the second trip is scheduled to depart using the same vehicle from the same stop at 11:20 AM. 
On Aug. 8, 2016, the arrival time at the last stop (Timepoint GRFSTATO) of the first trip (121359) is exceptionally late for 9 minutes, which contributes to the 10-minute departure delay at the beginning of the second trip. Since the scheduled layover time between the two trips is only 2 minutes (between 11:18 AM and 11:20 AM), any large delay at the first trip is very likely to transfer to the next trip. 
Therefore, the optimization problem should involve a process that considers not only the travel delay on segments, but also the improper lay over time between trips.

\begin{figure}[t]
\begin{center}
\centerline{\includegraphics[width=1.0\columnwidth]{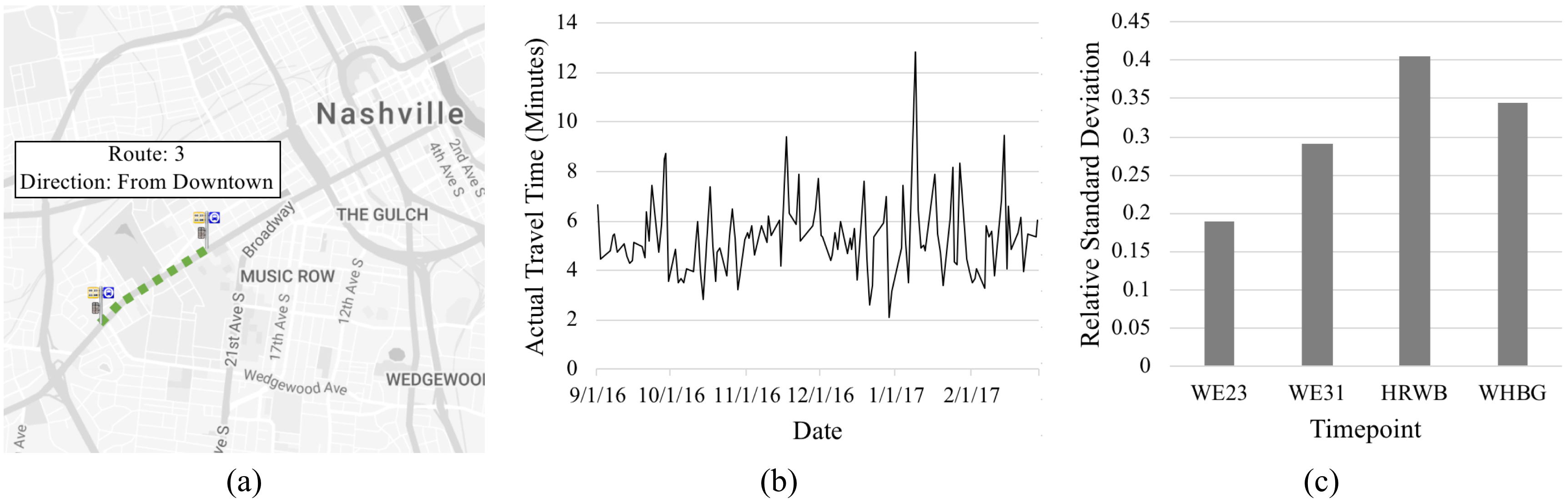}}
\caption{(a) A route segment on bus route 3 leaving downtown; (b) The variance of actual travel time and (c) the relative standard deviation of actual travel times on a bus route segment in time period between Sept. 1, 2016 and Feb. 28, 2017.}
\label{opt:fig:att_rsd}
\end{center}
\end{figure}

Figure~\ref{opt:fig:att_rsd} illustrates the large variation of bus travel time distribution. The example shows travel time data collected from bus trips depart at a specific time of the day on route 3 in Nashville. The coefficient of variation (also known as relative standard deviation)
, which is a standardized measure of dispersion of a probability distribution, is very high on all timepoints along the route. The complexity and uncertainty of travel times introduce great challenges to the task of timetable optimization.   

\subsection{Problem Definition}
\label{opt18:sec:formulation:problem}


For a given bus trip schedule $b$, let $H = \{h_{1},h_{2},...,h_{m}\}$ be a set of $m$ historical trips with each trip passing $n$ timepoints $\{s_{1},s_{2},...,s_{n}\}$. So the \textit{on-time performance} of the bus trip schedule $b$ can be defined as a ratio of an indicator function $I(h_{i}, s_{j})$ summed over all timepoints for all historical trips to the product of the total number of historical trips and total number of timepoints. The indicator function $I(h_i, s_j)$ is 1 if $d_{i,j} \in [t_{early}, t_{late}]$, otherwise 0, where $d_{i,j} = t^{arrival}_{h_{i}, s_{j}} - T^{arrival}_{h_{i}, s_{j}}$


The objective is to design a schedule optimization problem to generate new $T^{departure}_{h,s}$, ensuring on-time performance maximization.  $t_{early}$ and $t_{late}$ are two time parameters pre-defined by the transit authority as a measure of schedule maintenance and $d_{i,j}$ is the actual delay that arriving in timepoint $s_{j}$. 

\section{Data Store}
\label{opt18:sec:data}


\subsection{Data Sources}

We established a cloud data store and reliable transmission mechanisms to feed our 
Nashville Metropolitan Transit Authority (MTA) updates the bus schedule information every six months and provides the schedule to the public via GTFS files. In order to coordinate and track the actual bus operations along routes, MTA has deployed sensor devices at specially bus stops (called timepoints) to accurately record the arrival and departure times. In Nashville, there are over 2,700 bus stops all over the city and 573 of them are timepoint stops. City planners and MTA engineers analyze the arrival and departure records regularly to update the transit schedule. The details of the datasets are as follows:

\begin{itemize}
\item \textit{Static GTFS. } This dataset defines the static information of bus schedule and associated geographic information, such as routes, trips, stops, departure times, service days, etc. The dataset is provided in a standard transit schedule format called General Transit Feed Specification (GTFS).
\item \textit{GTFS-realtime.} This dataset is recorded real-time transit information in GTFS-realtime format, which include bus locations, trip updates and service alerts. The GTFS-realtime feed is collected and stored in one-minute interval.
\item \textit{Timepoints.} This dataset provides accurate and detailed historical arrival and departure records at timepoint stops. The information include route, trip, timepoint, direction, vehicle ID, operator, actual arrival and departure time, etc. The dataset is not available in real-time but collected manually by Nashville MTA at the end of each month.
\end{itemize}

Even though the same timepoint datasets are utilized in the study, the proposed method is not limited to the timepoint datasets and can use some surrogate data sources: (1) automatic passenger counters (APC) data: APC datasets records both passenger counts and departure/arrival times at stops (2) GTFS-realtime feed: the real-time bus locations reported by automatic vehicle locator (AVL) installed on buses. Compared with timepoint datasets, APC data also provides accurate times at normal stops thus it is the most suitable alternative dataset. However, GTFS-realtime suffers from low sampling rate and low accuracy in the city and may reduce the performance of the proposed mechanism.

\subsection{Data Cleaning}

Since raw transit dataset often contains missing, duplicate and erroneous samples, preprocessing is a necessary step to prepare a clean and high-quality dataset. 

Missing data issue occurs due to hardware or network problems. Generally, there are samples with missing data can be dropped or filled with a specific or average values. Duplicated data (e.g., a bus trip is recorded more than one time) will oversample certain delay values and make the delay dataset biased. We drop the trips with no historical records and remove duplicated records. 

Outliers are values that are distant from most of the observed data in presence of which clustering can be inappropriate. K-means clustering algorithm is also sensitive to outliers present in the data. The approach taken here is to calculate Median Absolute Deviation (MAD), a robust measure of statistical dispersion. The MAD of a data set [$X=(x_1, x_2,..., x_n)$] can be calculated as: $MAD = median( |x_i - median(X)| )$. For normal distribution the scaled MAD is defined as (MAD/0.6745), approximately equal to the standard deviation. $x_i$  is considered an outlier if the difference between $x_i$ and median is larger than three times of standard deviation (i.e. scaled MAD).

\section{Timetable Optimization Mechanisms}
\label{opt18:sec:solutions}




\subsection{Month Grouping by Clustering Analysis}
\label{opt18:sec:clustering}

This section introduces a clustering analysis mechanism that groups months with similar transit delay patterns together and the results will later be used to generate separate timetables for each group. 

\textit{Feature Engineering.} 
We assume the monthly delay patterns can be represented by the mean, median and standard deviation that derived from historical delay data. Considering a bus trip consists of $n$ timepoints, there are $n-1$ segments between the timepoints. The mean value $\mu$, the median value $m$, and the standard deviation $\sigma$ of the historical travel times for each timepoint segment in each month are integrated to generate feature vectors to represent the historical delay data distribution:
\begin{equation}
[\mu_1,m_1,\sigma_1,\mu_2,m_2,\sigma_2,...,\mu_{n-1},m_{n-1},\sigma_{n-1}]
\label{equa-distribution-a-month}
\end{equation}

\textit{Month Clustering.} Clustering is an unsupervised/supervised learning technique for grouping similar data. We employ k-means algorithms to identify the homogeneous groups where months share similar patterns. The trip data per month is first normalized and then clustered using feature vectors (in Equation~\ref{equa-distribution-a-month}) by K-Means algorithm:
\begin{equation}
\argmin\limits_{S} \sum_{i=1}^{k} \sum_{x \in S_{i}}\| x - \mu_{i} \|^{2}
\label{equa:k-means}
\end{equation}
where $\mu_{i}$ is the mean of all datapoints in cluster $S_{i}$. Determining the optimal number of clusters in a data set is a fundamental issue in partitioning clustering. For k-means algorithms, the number of clusters is a hyper-parameter that needs to be set manually. An upper bound is set in advance.
 Elbow \cite{kodinariya2013review}, 
Silhouette \cite{rousseeuw1987silhouettes} and gap statistic \cite{tibshirani2001estimating} methods are popular direct and statistical methods to find the optimal number of clusters. Particularly, Silhouette analysis is employed in this study to measure how close each point is to others within one cluster. The silhouette score $s(i)$ is defined as:
\begin{equation}
\begin{aligned}
s(i) = \frac{b(i) - a(i)}{max\{a(i), b(i)\}}
\end{aligned}
\label{eq:silhouete}
\end{equation}
where for each data point with index $i$ in the cluster, $a_{i}$ is the average distance 
between $data_i$ and the rest of data points in the same cluster, $b_{i}$ is the 
smallest average distance between $data_i$ and every other cluster. Some other clustering techniques that can be applied to these kind of problem can be found in \cite{8301693}.

\begin{figure}[t]
\vspace{-0.1in}
\begin{center}
\centerline{\includegraphics[scale=0.17]{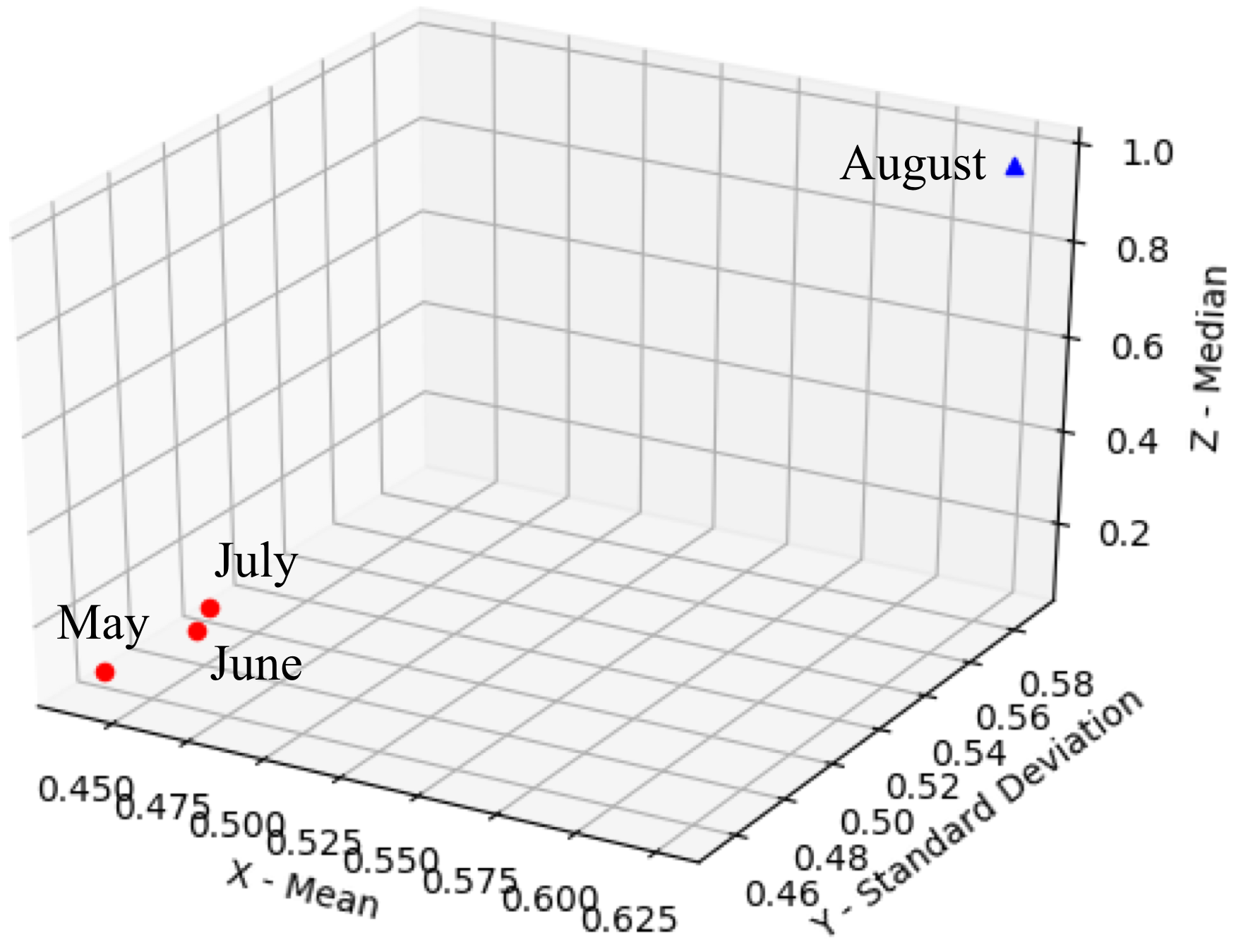}}
\caption{The feature vectors [mean, standard deviation, median] of the travel time in 4 months of 2016 for a segment (WE23-MCC5\_5) on a bus trip of route 5. }
\label{fig:month_cluster}
\end{center}
\vspace{-0.3in}
\end{figure}

\textit{Example.} Figure~\ref{fig:month_cluster} plots the [mean, standard deviation, median] vectors of the monthly travel time for a segment (WE23-MCC5\_5) on a bus trip of route 5 (Figure~\ref{fig:combinedfigs}(a)). From figure~\ref{fig:month_cluster} the monthly variation of the data is evident and hence two clusters ([May, June, July] and [August]) can be formed from these four months of data to prepare distinct schedules for the clusters.


\subsection{Estimating On-time Performance of Transit Schedules}
\label{opt18:sec:on-time-estimate}


\textit{Historical Dwell Time Estimation.} Travel demand at bus stops is important statistics for setting up proper schedule times. However, for bus systems without automatic passenger counters (APCs), historical travel demand (represented by number of commuters boarding) is not available in original datasets. To get demand patterns, we utilize historical arrival and departure times to estimate the dwell time caused by passengers. Particularly, we consider the following two scenarios in historical records: (1) when a bus turns up at a stop, earlier than scheduled time, the waiting time between the scheduled time and actual departure time is used, (2) on the other hand, when it turns up later than scheduled time, the waiting time between the actual arrival time and departure time is used. As shown in Table~\ref{opt18:tbl:example}, for the timepoint SY19 on trip 1 with scheduled time of 11:02 AM: 
\begin{itemize}
\item For the case when a bus arrived earlier at 10:58 AM instead of the scheduled time at 11:02 AM and departed at 11:04 AM, as the bus would always wait there at least for 4 minutes (the difference between the actual and scheduled arrival time) irrespective of presence of passengers, the dwell time caused by passengers is calculated as the additional time taken for departure after the scheduled time (11:04 AM - 11:02 AM = 2 minutes). 
\item On the other hand, if the bus arrived later at 11:05 AM and departed at 11:06 AM, then the dwell time caused by passengers is calculated as the additional time spent after the actual arrival time (11:06 AM - 11:05 AM = 1 minutes).
\end{itemize}

\textit{Arrival Time Estimation.} The arrival time of a bus at a stop is impacted by two factors: (1) travel times at segments before the stop, and (2) dwell times at the previous stops. We assume that a bus will wait until the scheduled time if it arrives earlier than the scheduled time, and the historical travel time between two timepoints will remain the same in the simulation. In order to obtain an estimate of the arrival time, the historical dwell time caused by commuters (which in turn is representative of the historical travel demand), is factored into account by adding it to the arrival time at any timepoint. The simulation will stall for an additional time till the new scheduled time is reached in the event that the previous sum is earlier than the new scheduled time. By taking into consideration the simulated departure time $st^{depart}_{h, s_{j}}$ at previous timepoint $s_j$, the actual travel time $t^{arrive}_{s_{j+1}}-t^{depart}_{s_j}$ between $s_j$ and $s_{j+1}$, the dwell time $t^{dwell}_{s_{j+1}}$, the simulated departure time $st^{depart}_{h, s_{j+1}}$ at a timepoint $s_{j+1}$ can be found out. The new schedule time $T^{depart}_{h, s{s_{j+1}}}$ at $s_{j+1}$ is expressed as: 
\begin{equation}
st^{depart}_{h, s_{j+1}} = \max(T^{depart}_{h, s{s_{j+1}}}, st^{depart}_{h, s_{j}}+(t^{arrive}_{s_{j+1}}-t^{depart}_{s_j})+t^{dwell}_{s_{j+1}})
\end{equation}

\subsection{Timetable Optimization Using a Greedy Algorithm}
\label{opt18:subsec:greedy}





We employed a  greedy algorithm that adjusts the scheduled arrival time greedily and sequentially for the succeeding segments between timepoints. The main objective is to optimize the bus arrival time for succeeding timepoints such that new optimized schedule is guaranteed to maximize the probability of bus arrivals between any two consecutive stops with delay bounded in the desired range of [$t_{early}, t_{late}$]. 


We utilized the empirical cumulative distribution function (CDF) to evaluate the percentage of historical delay in desired range instead of assuming that the data is drawn from any specific distribution (e.g. Gaussian distribution).

An empirical CDF is a non-parametric estimator of the CDF of a random variable. The empirical CDF of variable $x$ is defined as:
\begin{equation}
\hat{F}_n(x) = \hat{P}_n(X\leq x) = n^{-1} \sum_{n=1}^{n} I(x_i \leq x)
\label{equa:ecdf}
\end{equation}
where $I()$ is an indicator function:
\begin{equation}
I(x_i \leq x) = 
\begin{cases}
1, & \text{if}\ x_i \leq x \\
0, & \text{otherwise}
\end{cases}
\label{equa:ecdf-indicator}
\end{equation}
Then the CDF of $x$ in range [$x+t_{early},x+t_{late}$] can be calculated using the following equation:
\begin{equation}
\begin{aligned}
\hat{F}_n(x+t_{late})-\hat{F}_n(x+t_{early})       \\
= n^{-1} \sum_{n=1}^{n} I(x+t_{early} \leq x_i \leq x+t_{late})
\label{equa:ecdf-range}
\end{aligned}
\end{equation}



\subsection{Timetable Optimization Using Heuristic Algorithms}


The performance optimization for scheduling transit vehicles is a multidimensional problem and as such the objective function is nonconvex in nature consisting of several troughs and ridges. Hence, to compute the optimally scheduled routing strategy with acceptable time constraints, an approach powered by high quality of solution estimation techniques such as evolutionary algorithms and metaheuristics can be considered. 


\subsubsection{Genetic Algorithm}

Genetic algorithm \cite{Goldberg:1989:GAS:534133} is a heuristic optimization algorithm that derives from biology. The basic steps involved in genetic algorithms include initialization, selection, crossover, mutation, and termination. The timetable for each trip is decided by the scheduled departure time at the first stop as well as the scheduled travel time between any two subsequent timepoints along the trip. Since our goal is to update timetables to make the bus arrivals more on time, we assign the scheduled travel times between timepoints as chromosomes in populations, and use the on-time performance estimation mechanism proposed in Section~\ref{opt18:sec:on-time-estimate} as objective functions. The chromosome of the individual solutions in the genetic algorithm is a vector of integers representing travel time between subsequent timepoints. In order to reduce the search space and match the real-world scenarios, the travel time in each individual is re-sampled to a multiple of 60 seconds and restricted to the unit of minutes. The performance of this algorithm is governed by different hyperparameters such as population size, crossover and mutation rate controlling the algorithm's exploitation and exploration capability. The choice of such hyperparameters are explained in detail in section \ref{opt18:sec:evaluations}.

\subsubsection{Particle Swarm Optimization}

Eberhert and Kennedy \cite{488968} proposed particle swarm optimization (PSO) as a stochastic population based optimization algorithm which can work with non-differentiable objective function without explicitly assuming its underlying gradient disparate from gradient descent techniques. The interested reader is directed to \cite{psoreview} by Sengupta et al. for a detailed understanding of the algorithm. PSO has been shown to satisfactorily provide  solutions to a wide array of complex real-life engineering problems, usually out of scope of deterministic algorithms  \cite{7060145}\cite{BOUYER2018172}\cite{Banks2008}.  PSO exploits the collective intelligence arising out of grouping behavior of flocks of birds or schools of fish.This manifestation of grouping is termed as 'emergence', a phenomenon in which a cohort of individuals from a social network is aimed to accomplish a task beyond their individual capability. Likewise, each particle in the swarm, represents a potential solution to the multi-dimensional problem to be optimized. 

\textbf{Initialization} Each particle has certain position which can be thought of as a collection of co-ordinates representing the particle's existence in a specific region in the multidimensional hyperspace. As a particle is a potential solution to the problem, the particle's position vector has the same dimensionality as the problem. The velocity associated with each particle is the measure of the step size and the direction it should move in the next iteration.

Each particle in the swarm maintains an n-dimensional vector of travel times. At first, the position for each particle in the population is initialized with the set of travel time between the timepoints randomly selected between the minimum and maximum of the aggregated actual historical data. With swarm size as \textit{p}, every particle \textit{i} (1$<$\textit{i}$<$\textit{p}) maintains a position vector $x_{i}$=($x_{i1}$,$x_{i2}$,$x_{i3}$,...,$x_{in}$) and a velocity vector $v_{i}$=($v_{i1}$,$v_{i2}$,$v_{i3}$,...,$v_{in}$) and a set of personal bests $p_{i}$=($p_{i1}$,$p_{i2}$,$p_{i3}$,...,$p_{in}$).

\textbf{Optimization} At each iteration, the position of a particle is updated, and compared with the personal best (\textit{pbest}) obtained so far. If the fitness due to the position obained at current iteration is more (as it is a fitness maximization problem) than the \textit{pbest} obtained upto the previous iteration, then the current position becomes the personal best or \textit{pbest}, otherwise \textit{pbest} remains unchanged. Thus the best position of a particle obtained so far is stored as \textit{pbest}. The global best or \textit{gbest} is updated when the population's overall current best, i.e., the best of the \textit{pbsest}s is better than that found in the previous iteration.

After initializing positions and velocities, each particle updates its velocity based on previous velocity component weighted by an inertial factor , along with a component proportional to the difference between its current position and \textit{pbest} weighted by a cognition acceleration coefficient, and another component proportional to the difference between its current position and  (\textit{gbest}), weighted by a social acceleration coefficient. This is socio-cognitive model of PSO and facilitates information exchange between members of the swarm. Since all members are free to interact with each other, the flow of information is unrestricted and the PSO algorithm is said to have a 'fully-connected' topology. While updating the velocity,  a particle's reliance on its own personal best is dictated by its cognitive ability, and the reliance on the entire swarm's best solution is dictated by its social interactive nature. Hence those factors in the velocity component are weighted by the cognition acceleration coefficient \textit{c1} and social acceleration coefficient \textit{c2}. The new positions of the particles are updated as the vector sum of the previous positions and the current velocities. Thus the positions of the particles, are updated aiming towards intelligent exploration of the search space, and subsequent exploitation of the promising regions in order to find the optimal solution based on fitness optimization of the stated problem.

 After each iteration is completed, the velocity and position of a particle are updated as follows:
\begin{equation}
v_{i,j}(t+1)=w.v_{i,j}(t)+c_{1}.r_{1}(t).(p_{i,j}(t)-x_{i,j}(t))+c_{2}.r_{2}(t).(p_{g,j}(t)-x_{i,j}(t))
\end{equation}
\begin{equation}
x_{i,j}(t+1)=x_{i,j}(t)+v_{i,j}(t+1)
\end{equation}

$v_{i,j}$ and $x_{i,j}$ represent the velocity and position of the \textit{i-th} particle in the \textit{j-th} dimension. Cognition and social acceleration coefficients are indicated by $c_{1}$ and $c_{2}$, whereas $r_{1}$ and $r_{2}$ are random numbers uniformly distributed between 0 to 1. $p_{i,j}$ represents a particle$'$s personal best and $p_{g,j}$ represents the global best of the population. \textit{w} acts as an inertial weight factor controlling the exploration and exploitation of new positions in the search space and t denotes the number of iterations.

The problem is formulated as fitness maximization problem in order to bring out optimal travel times to improve on-time performance. Hence the personal best of a particle is updated as follows at the end of each iteration. 

\begin{equation}
p_{i,j}(t+1) = 
\begin{cases}
p_{i,j}(t), & \text{if}\ fitness(x_{i,j}(t+1)) \leq fitness(p_{i,j}(t))\\
x_{i,j}(t+1), & \text{if}\ fitness(x_{i,j}(t+1)) > fitness(p_{i,j}(t))\\
\end{cases}
\end{equation}


\begin{algorithm}[t]
\small
\vspace{-0.0in}
	\KwData{$D \leftarrow$ Historical timepoint datasets}
	\SetKwInOut{Input}{Input}
    \SetKwInOut{Output}{Output}
    \Input{
    (1) [$t_{early}$,$t_{late}$] $\leftarrow$ on-time range , (2) $maxIter$ $\leftarrow$ maximum number of iterations $maxIter$, (3) $npop$ $\leftarrow$ number of particles in the population size $npop$,  (4) $w$ $\leftarrow$ inertia weight, (5) $c1$ $\leftarrow$ cognition acceleration coefficient, (6) $c2$ $\leftarrow$ social acceleration coefficient, (7) $h$ $\leftarrow$ bus trip for optimization, (8) $upperLimit$ $\leftarrow$ upper limit of the number of clusters }
    \Output{Optimized schedule b at timepoints for bus trip $h$}

GetAllTimepoints($D$, $h$)\;
GetHistoricalData($D$, $h$)\;
$monthClusters \leftarrow$ ClusterMonthData($upperLimit$)\;
\For{monthCluster $\in$ $monthClusters$}{
$P \leftarrow$ []\;
\For{population size $npop$}{
Initialize each particle with random position and velocity

	$P \leftarrow P \cup InitialIndividual()$\;
}

\While{$maxIter$ is not reached}{
	Evaluate the fitness function (\textit{J}) for each particle's position (\textit{x})
    
    if \textit{J(x)} > \textit{J(pbest)}, then \textit{pbest = x}
    
    $gbest \leftarrow$ Update if the population's overall current best is better than that in previous iteration
    
    Update the velocity of each particle according to equation \textit{(10)}
    
    Update the position of each particle according to equation \textit{(11)}
    
}
Give \textit{gbest} as the optimal schedule b at timepoints for bus trip \textit{h}
}
 
 \caption{Particle Swarm Optimization algorithm for bus on-time performance optimization}
\label{algorithm:particle swarm optimization}
\end{algorithm}

\textbf{Termination}
The termination condition set for PSO is the predefined maximum number of iterations. Since the optimized on time performance is different for each trip, the termination condition is not set as any predefined upper limit of the fitness value. With other hyperparameters fixed PSO can produce the optimal solution approximately in 30 iterations for this problem.

The pseudo code for PSO is discussed in Algorithm 3.  Historical timepoint datasets
are used to conduct the particle swarm optimization algorithm for this problem. The input includes on-time range, maximum number of
iterations, number of particles in the population size, inertia factor, cognition and social acceleration coefficient,
bus trip and upper limit of number of month clusters.

\section{Evaluation of the Results}
\label{opt18:sec:evaluations}


\subsection{Evaluating the Clustering Analysis}
\label{opt18:sec:sensitivity}

 To evaluate the effectiveness of the clustering analysis, we compared the optimized on-time performance with and without a clustering analysis step: (1) months are not clustered and a single timetable is generated for all months, (2) month clustering is conducted at first and the optimization algorithms is applied on different month groups to generate separate timetables.

 Table~\ref{opt18:tbl:exp1} shows the original and optimized on-time performance on average across all bus routes. Using the genetic algorithm without clustering step improved the original performance from 57.79\% to 66.24\%. By adding the clustering step which groups months with similar patterns the performance was improved to 68.34\%.

\begin{table}[tbp]
\centering
\small
\caption{Comparison of original and optimized on-time performance averaged across all bus routes for GA without and with clustering and PSO with clustering respectively. 
}
\begin{tabular}{|c|c|c|c|c|}
\hline
 & Original  & GA w/o. Clustering & GA w. Clustering
 & PSO w. Clustering 
 \\
\hline
On-time Perf. & 57.79\% & 66.24\% & 68.34\% 
& 68.93\% 
\\   
\hline
\end{tabular}
\label{opt18:tbl:exp1}
\end{table}

\begin{figure}[tphb]
\begin{center}
\centerline{\includegraphics[scale=0.35]{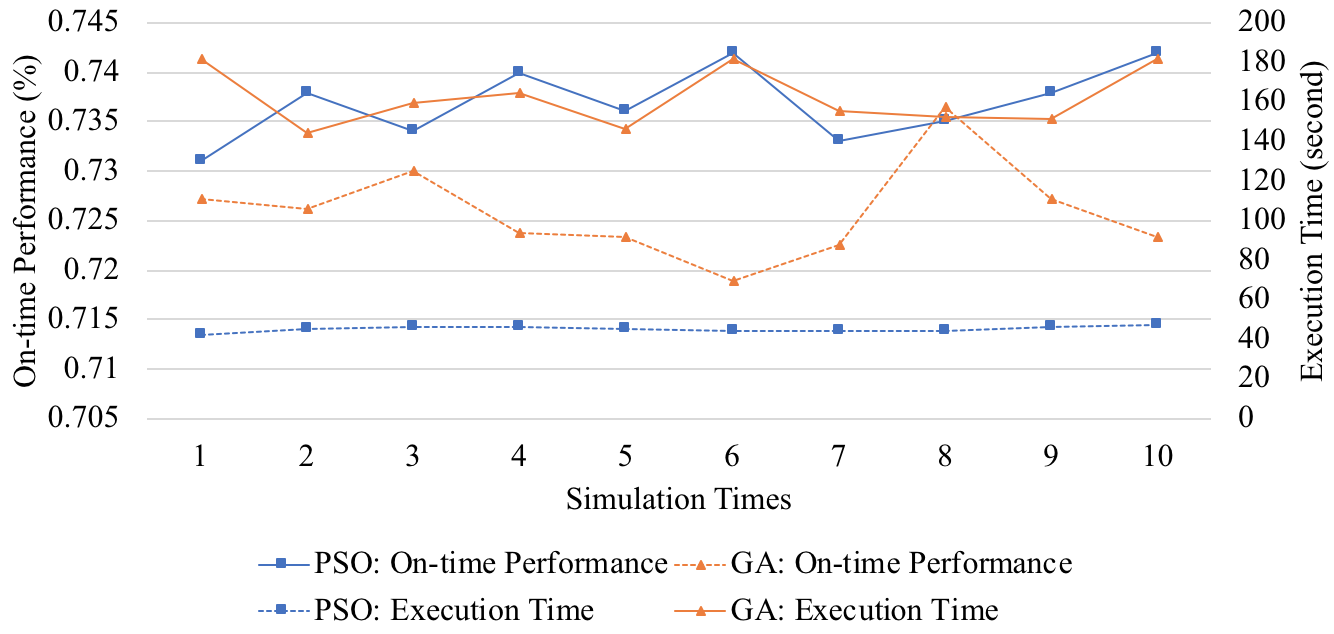}}
\caption{The chart shows the simulation results of on-time performance and execution times for GA and PSO to run 10 times.}
\label{opt18:fig:stability}
\end{center}
\end{figure}

\subsection{Comparing Optimization Performance of Greedy, Genetic and PSO Algorithms}



\begin{figure}[tphb]
\begin{center}
\centerline{\includegraphics [scale =0.2]{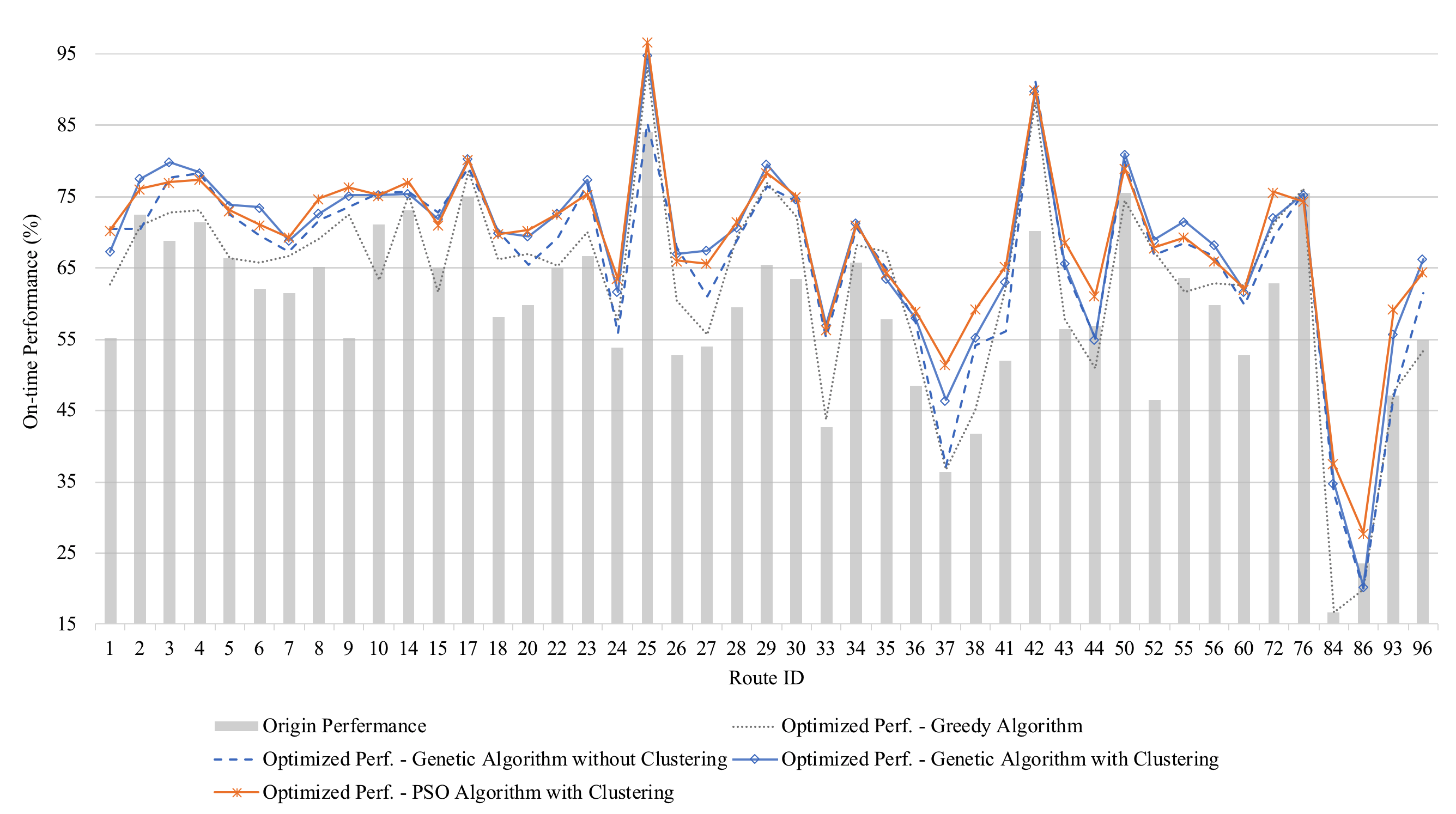}}
\caption{The original on-time performance and the optimized on-time performance using greedy algorithm, genetic algorithm with/without clustering analysis and PSO algorithm.}
\label{opt18:fig:on-time}
\end{center}
\end{figure}

The original on-time performance, optimized on-time performance using greedy algorithm, genetic algorithm and PSO are illustrated in Figure~\ref{opt18:fig:on-time}. It is observed that while all the algorithms can improve the on-time performance, the genetic algorithm and PSO outperforms the greedy algorithm because they optimize the schedule for all timepoint segments on each trip all together. The original on-time performance of all bus routes  is 57.79\%. The greedy algorithm improved it to 61.42\% and the genetic algorithm improved it further to 68.34\%. The PSO algorithm has a slightly better optimized on-time performance of 68.93\%. Figure~\ref{opt18:fig:stability} shows the simulation results of the stability analysis for GA and PSO. Even though GA and PSO got similar on-time performance, with PSO surpassing the performance of GA by a small extent, the execution times of PSO are much less than GA and are more stable.


\subsection{Sensitivity Analysis on the Hyper-parameters of the Genetic algorithm}
\label{opt18:sec:sensitivity:results}



We designed three simulations that choose different hyper-parameters: (1) population sizes that range from 10 to 110, (2) crossover rates that range from 0.1 to 1.0, (3) mutation rates that range from 0.1 to 1.0. Real-world data is collected from Route 5, which is one of major bus routes that connects downtown Nashville and the southwest communities in Nashville. The route contains 6 timepoint stops and 5 segments between the 6 timepoint stops. The bus trips with direction from Downtown are selected. The goal is to maximize the on-time performance for these trips by optimizing the schedule time at the 6 timepoint stops.

\begin{figure}[tphb]
\begin{center}
\centerline{\includegraphics[scale = 0.62]{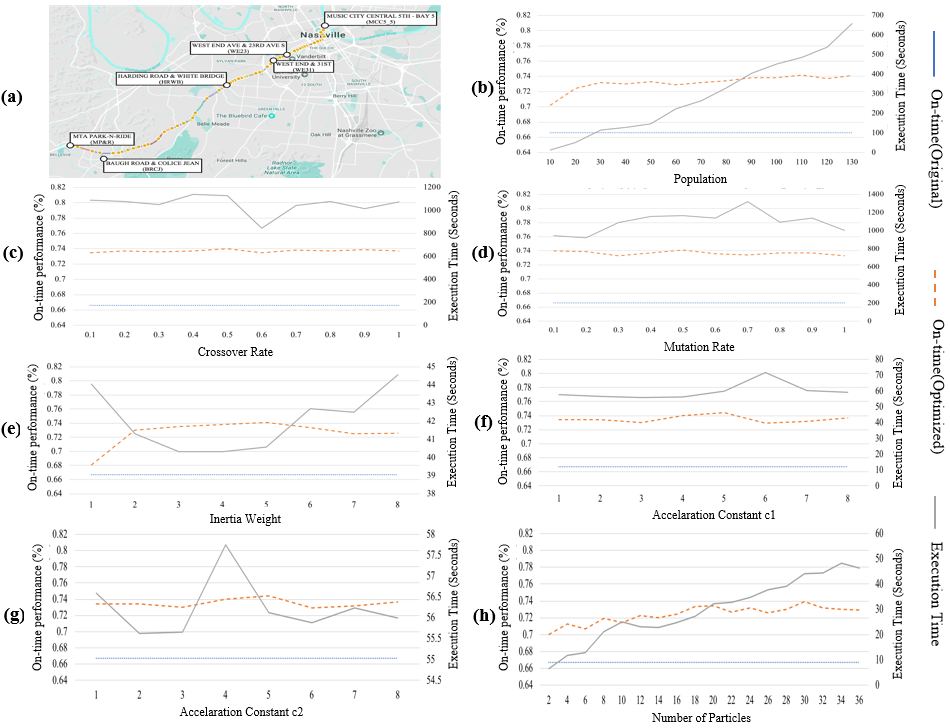}}
\caption{(a) Timepoints on bus route 5 in Nashville \cite{sun2017unsupervised}, (b) The  the on-time performance and overall execution time for different population sizes for GA, (c) The on-time performance and overall execution time for different crossover rates, which controls the exploitation ability of the GA, (d) The  on-time performance and overall execution time for different mutation rates, which controls the exploration ability of the GA, (e)  The on-time performance and overall execution time for different inertia weights, exploring new regions of search space in PSO, (f) The on-time performance and overall execution time for different cognition acceleration coefficients c1, in PSO, (g) The on-time performance and overall execution time for different social acceleration coefficients c2, in PSO, (h) The on-time performance and overall execution time for various population size, in PSO}
\label{fig:combinedfigs}
\end{center}
\end{figure}


Figure~\ref{fig:combinedfigs}(b) shows the simulation results of choosing different population sizes. Increasing the population size from 10 to 90 results a better on-time performance, however, increasing the size ever further doesn't help making the on-time performance any better. On the other hand, the total time increases linearly as the population size grows. So a population size around 90 is the optimal size to use. 


Figure~\ref{fig:combinedfigs}(c) illustrates results of using different crossover rates. The optimized on-time performance remains almost the same for the crossover range, but there is a significant difference in terms of the total execution time. The crossover rate impacts the exploitation ability. A proper crossover rate in the middle of the range can faster the process to concentrate on an optimal point. 


Figure~\ref{fig:combinedfigs}(d) show the simulation results when using different mutation rates. The total execution time is small when the mutation rate is either very small or very large. Mutation rates controls the exploration ability. During the optimization, a small mutation rate will make sure the best individuals in a population do not vary too much in the next iteration and thus is faster to get stable around the optimal points. So we suggest setting a very small mutation rates when running the proposed algorithm.

\subsection{Sensitivity Analysis on the Hyper-parameters of Particle Swarm Optimization}


We designed four simulation setups that choose different hyper-parameters: (1) The inertial weight factor,\textit{w} that range from 1 to 8, (2) Social acceleration coefficient \textit{c1} that range from 1 to 8, (3) Cognition acceleration coefficient \textit{c2} that range from 1 to 8, and (4) Number of particles that range from 2 to 36. Real-world data regarding bus timings is collected from Route 8, which is one of the major bus routes that connects Music City Central Nashville and the Lipscomb University in Nashville. The route contains 5 timepoint stops and 4 segments between the 5 timepoint stops. The goal is to maximize the on-time performance for these trips by optimizing the schedule time at the 5 timepoint stops.

Figure~\ref{fig:combinedfigs}(e) shows the simulation result while optimizing for the inertial weight \textit{w} by varying it. It is observed that the optimized on-time performance is at its peak when \textit{w} is nearly equal to 5 with less execution time. Performance deteriorates along with an increase in execution time as the selection is moved away from 5. So, an optimal value to choose for w, will be somewhere around 5.

Figure~\ref{fig:combinedfigs}(f) shows the simulation result for optimizing the cognition acceleration coefficient \textit{c1} by varying it. The particle has a velocity component towards its own best position weighted by \textit{c1}, hence the term 'cognitive'. It is observed that the optimized on-time performance is improved when \textit{c1} increases from 3 to 5 with less execution time. After that the performance deteriorates along with increase in execution time. So, an optimal value to choose for \textit{c1}, will be within the range specified.


Figure~\ref{fig:combinedfigs}(g) shows the simulation result for optimizing for the social acceleration coefficient \textit{c2} by varying it. The particle has a velocity component towards the global best position weighted by \textit{c2}, hence the term social. It is observed that the optimized on-time performance is improved when \textit{c2} is equal to 5 with less execution time. Also, \textit{c2} being 4 produces good results, but there is an increase in execution time at that value. But the overall effect of parameter \textit{c2} affects the on-time performance only within a range of two percent. Sometimes, PSO is able to produce optimal or near optimal performance, when all other hyperparameters are fixed, and thus is not sensitive to a particular hyperparameter, which is the case considered here. So an optimal value to choose for \textit{c2}, may be close to 5, maintaining approximately a ratio near to 1:1:1 among \textit{w}, \textit{c1} and \textit{c2}. 


Figure~\ref{fig:combinedfigs}(h) shows the simulation result for optimizing the number of particles by varying the population size. It is observed that the optimized on-time performance is maximized when the number of particles reaches 30. The execution time increases with the number of particles, so it is better to choose such number of particles that produces the best pair in the accuracy-execution time tradeoff. So, the population size can be chosen as 30 in this case as it yields equally efficient results with a relatively small execution time.




Although a good insight about choice of hyperparameters can be obtained from this sensitivity analysis, variations of the hyperparameters may produce better results in specific routes.

\section{Conclusion}
\label{opt18:sec:conclusion}

In this paper, we presented research findings within a bus on-time performance optimization framework  that significantly extends our prior work \cite{sun2017unsupervised} by proposing a stochastic optimization toolchain and presenting sensitivity analyses on choosing optimal hyper-parameters. Particularly, we describe an unsupervised analysis mechanism to find out how months with similar delay patterns can be clustered to generate new timetables. A classical, fully-connected PSO is benchmarked against a greedy algorithm as well as a genetic algorithm in order to optimize the schedule time to maximize the probability of bus trips that reach the desired on-time range. It is observed that the PSO implementation improves the bus on-time performance compared to other heuristics while requiring lesser execution time across all routes. Simulations of optimization performance as well as sensitivity analyses on the hyper-parameters of the GA and PSO algorithms are conducted. The results indicate different strategies for choosing between the genetic algorithm and PSO, and selecting optimal hyper-parameters guided by the problem specificity and resource availability. With the knowledge of this extensive study on applying guided random search techniques for bus on-time performance optimization and the selection of hyperparameters that generate promising results, a possible extension of this generalizable architecture to other real-world optimization problems is worth looking at as future work. 

\section*{Acknowledgments}
This work is supported by The National Science Foundation under the award numbers CNS-1528799 and CNS-1647015 and 1818901 and a TIPS grant from Vanderbilt University.  We acknowledge the support provided by our partners from Nashville Metropolitan Transport Authority.



\bibliographystyle{splncs04}
\bibliography{mybibliography}

\begin{thebibliography}{10}
\providecommand{\url}[1]{\texttt{#1}}
\providecommand{\urlprefix}{URL }
\providecommand{\doi}[1]{https://doi.org/#1}

\bibitem{arhin2014bus}
Arhin, S.A., Noel, E.C., Dairo, O.: Bus stop on-time arrival performance and
  criteria in a dense urban area. International Journal of Traffic and
  Transportation Engineering  \textbf{3}(6),  233--238 (2014)

\bibitem{apta2017report}
Association, A.P.T.: Ridership report archives  (2017)

\bibitem{Banks2008}
Banks, A., Vincent, J., Anyakoha, C.: A review of particle swarm optimization.
  part ii: hybridisation, combinatorial, multicriteria and constrained
  optimization, and indicative applications. Natural Computing  \textbf{7}(1),
  109--124 (Mar 2008). \doi{10.1007/s11047-007-9050-z},
  \url{https://doi.org/10.1007/s11047-007-9050-z}

\bibitem{benn1995bus}
Benn, H.: Bus route evaluation standards, transit cooperative research program,
  synthesis of transit practice 10. Transportation Research Board, Washington,
  DC  (1995)

\bibitem{BOUYER2018172}
Bouyer, A., Hatamlou, A.: An efficient hybrid clustering method based on
  improved cuckoo optimization and modified particle swarm optimization
  algorithms. Applied Soft Computing  \textbf{67},  172 -- 182 (2018).
  \doi{https://doi.org/10.1016/j.asoc.2018.03.011},
  \url{http://www.sciencedirect.com/science/article/pii/S1568494618301273}

\bibitem{chakroborty2003genetic}
Chakroborty, P.: Genetic algorithms for optimal urban transit network design.
  Computer-Aided Civil and Infrastructure Engineering  \textbf{18}(3),
  184--200 (2003)

\bibitem{chakroborty1995optimal}
Chakroborty, P., Deb, K., Subrahmanyam, P.: Optimal scheduling of urban transit
  systems using genetic algorithms. Journal of transportation Engineering
  \textbf{121}(6),  544--553 (1995)

\bibitem{7060145}
Dhabal, S., Sengupta, S.: Efficient design of high pass fir filter using
  quantum-behaved particle swarm optimization with weighted mean best position.
  In: Proceedings of the 2015 Third International Conference on Computer,
  Communication, Control and Information Technology (C3IT). pp.~1--6 (Feb
  2015). \doi{10.1109/C3IT.2015.7060145}

\bibitem{eranki2004model}
Eranki, A.: A model to create bus timetables to attain maximum synchronization
  considering waiting times at transfer stops  (2004)

\bibitem{fan2006optimal}
Fan, W., Machemehl, R.B.: Optimal transit route network design problem with
  variable transit demand: genetic algorithm approach. Journal of
  transportation engineering  \textbf{132}(1),  40--51 (2006)

\bibitem{Goldberg:1989:GAS:534133}
Goldberg, D.E.: Genetic Algorithms in Search, Optimization and Machine
  Learning. Addison-Wesley Longman Publishing Co., Inc., Boston, MA, USA, 1st
  edn. (1989)

\bibitem{hairong2009optimal}
Hairong, Y., Dayong, L.: Optimal regional bus timetables using improved genetic
  algorithm. In: Intelligent Computation Technology and Automation, 2009.
  ICICTA'09. Second International Conference on. vol.~3, pp. 213--216. IEEE
  (2009)

\bibitem{ibarra2012synchronization}
Ibarra-Rojas, O.J., Rios-Solis, Y.A.: Synchronization of bus timetabling.
  Transportation Research Part B: Methodological  \textbf{46}(5),  599--614
  (2012)

\bibitem{488968}
Kennedy, J., Eberhart, R.: Particle swarm optimization. In: Neural Networks,
  1995. Proceedings., IEEE International Conference on. vol.~4, pp. 1942--1948
  vol.4 (Nov 1995). \doi{10.1109/ICNN.1995.488968}

\bibitem{10.1007/978-3-319-31753-3_44}
Khiari, J., Moreira-Matias, L., Cerqueira, V., Cats, O.: Automated setting of
  bus schedule coverage using unsupervised machine learning. In: Bailey, J.,
  Khan, L., Washio, T., Dobbie, G., Huang, J.Z., Wang, R. (eds.) Advances in
  Knowledge Discovery and Data Mining. pp. 552--564. Springer International
  Publishing, Cham (2016)

\bibitem{kodinariya2013review}
Kodinariya, T.M., Makwana, P.R.: Review on determining number of cluster in
  k-means clustering. International Journal  \textbf{1}(6),  90--95 (2013)

\bibitem{nayeem2014transit}
Nayeem, M.A., Rahman, M.K., Rahman, M.S.: Transit network design by genetic
  algorithm with elitism. Transportation Research Part C: Emerging Technologies
   \textbf{46},  30--45 (2014)

\bibitem{neff20172016}
Neff, J., Dickens, M.: 2016 public transportation fact book  (2017)

\bibitem{rousseeuw1987silhouettes}
Rousseeuw, P.J.: Silhouettes: a graphical aid to the interpretation and
  validation of cluster analysis. Journal of computational and applied
  mathematics  \textbf{20},  53--65 (1987)

\bibitem{8301693}
{Sengupta}, S., {Basak}, S., {Peters}, R.A.: Data clustering using a hybrid of
  fuzzy c-means and quantum-behaved particle swarm optimization. In: 2018 IEEE
  8th Annual Computing and Communication Workshop and Conference (CCWC). pp.
  137--142 (Jan 2018). \doi{10.1109/CCWC.2018.8301693}

\bibitem{psoreview}
Sengupta, S., Basak, S., Peters, R.A.: Particle swarm optimization: A survey of
  historical and recent developments with hybridization perspectives. Machine
  Learning and Knowledge Extraction  \textbf{1}(1),  157--191 (2018),
  \url{http://www.mdpi.com/2504-4990/1/1/10}

\bibitem{DBLP:journals/corr/abs-1905-13294}
Sengupta, S., Basak, S., Saikia, P., Paul, S., Tsalavoutis, V., Atiah, F.,
  Ravi, V., Peters, R.A.: A review of deep learning with special emphasis on
  architectures, applications and recent trends. CoRR  \textbf{abs/1905.13294}
  (2019), \url{http://arxiv.org/abs/1905.13294}

\bibitem{sun2017unsupervised}
Sun, F., Samal, C., White, J., Dubey, A.: Unsupervised mechanisms for
  optimizing on-time performance of fixed schedule transit vehicles. In: Smart
  Computing (SMARTCOMP), 2017 IEEE International Conference on. pp.~1--8. IEEE
  (2017)

\bibitem{szeto2011simultaneous}
Szeto, W.Y., Wu, Y.: A simultaneous bus route design and frequency setting
  problem for tin shui wai, hong kong. European Journal of Operational Research
   \textbf{209}(2),  141--155 (2011)

\bibitem{tibshirani2001estimating}
Tibshirani, R., Walther, G., Hastie, T.: Estimating the number of clusters in a
  data set via the gap statistic. Journal of the Royal Statistical Society:
  Series B (Statistical Methodology)  \textbf{63}(2),  411--423 (2001)

\bibitem{ting2005schedule}
Ting, C.J., Schonfeld, P.: Schedule coordination in a multiple hub transit
  network. Journal of urban planning and development  \textbf{131}(2),
  112--124 (2005)

\bibitem{wang2017data}
Wang, Y., Zhang, D., Hu, L., Yang, Y., Lee, L.H.: A data-driven and optimal bus
  scheduling model with time-dependent traffic and demand. IEEE Transactions on
  Intelligent Transportation Systems  \textbf{18}(9),  2443--2452 (2017)

\bibitem{wu2016multi}
Wu, Y., Yang, H., Tang, J., Yu, Y.: Multi-objective re-synchronizing of bus
  timetable: Model, complexity and solution. Transportation Research Part C:
  Emerging Technologies  \textbf{67},  149--168 (2016)

\bibitem{psotraffic}
Zhong, S., Zhou, L., Ma, S., Jia, N., Zhang, L., Yao, B.: The optimization of
  bus rapid transit route based on an improved particle swarm optimization.
  Transportation Letters  \textbf{10}(5),  257--268 (2018).
  \doi{10.1080/19427867.2016.1258972},
  \url{https://doi.org/10.1080/19427867.2016.1258972}

\end{thebibliography}
\end{document}